\RequirePackage{fix-cm}
\documentclass[twocolumn]{svjour3}          
\smartqed  
\usepackage{graphicx}
\usepackage{subfigure}
\usepackage{amsmath}
\usepackage{mathptmx}

\usepackage[dvipsnames]{xcolor}
\newcommand{\ch}[1]{\textcolor[rgb]{0,0,0}{#1}}
\journalname{The International Journal of Advanced Manufacturing Technology}
\begin{document}

\title{A Physics-Informed Machine Learning Model for Porosity Analysis in Laser Powder Bed Fusion Additive Manufacturing}

\author{Rui Liu$^{1}$ \and Sen Liu$^{2,3}$ \and Xiaoli Zhang$^{2, 3^*}$}

\institute{
Rui Liu is the first author (rliu11@kent.edu), Xiaoli Zhang is the corresponding author (xlzhang@mines.edu).
\at $^{1}$ Cognitive Robotics and AI Lab (CRAI), College of Aeronautics and Engineering, Kent State University, Kent, OH 44240
\at $^{2}$ Alliance for the Development of Additive Processing Technologies, Colorado School of Mines, Golden, CO 80401
\at $^{3}$ Department of Mechanical Engineering, Colorado School of Mines, Golden, CO 80401
}

\date{Received: date / Accepted: date}

\maketitle

\begin{abstract}
To control part quality, it is critical to analyze pore generation mechanisms, laying theoretical foundation for future porosity control. Current porosity analysis models use machine setting parameters, such as laser angle and part pose. However, these setting-based models are machine dependent, hence they often do not transfer to analysis of porosity for a different machine. To address the first problem, a physics-informed, data-driven model (PIM), which instead of directly using machine setting parameters to predict porosity levels of printed parts, it first interprets machine settings into physical effects, such as laser energy density and laser radiation pressure. Then, these physical, machine independent effects are used to predict porosity levels according to “pass”, “flag”, “fail” categories instead of focusing on quantitative pore size prediction. With six learning methods’ evaluation, PIM proved to achieve good performances with prediction error of 10$\sim$26\%. Finally, pore-encouraging influence and pore-suppressing influence were analyzed for quality analysis.

\end{abstract}

\section{Introduction}
\label{intro}
Metal additive manufacturing (metal AM) groups together laser sintering (LS), selective laser melting (SLM), direct metal deposition (DMD), and laser engineered net shape (LENS); processes that utilize a laser heat source to join metal materials for producing mechanical parts directly from 3D models in an incremental manner \cite{ref1}\cite{ref2}. Recently, metal AM has been implemented on manufacturing aerospace products such as jet engines and body shells \cite{ref3}, automotive products such as car gearboxes and seats \cite{ref2}, medical devices such as micro-forceps, and tissue approximation devices \cite{ref4}. Laser Additive Manufacturing (LAM) brings benefits, such as adopting complex product structures, reducing production time cycle, enabling rapid design and testing of conceptual prototypes, simplifying tedious producing procedures, and achieving extraordinary mechanical system compactness \cite{ref2}\cite{ref5}\cite{ref6}. \ch{Because of the short laser-material interaction times and highly localized heat input, the thermal gradients and rapid solidification lead to a build-up of thermal stresses and non-equilibrium phases \cite{ref7}\cite{ref8}. The non-optimal process parameters can cause melt pool instabilities, which leads to porosity or geometric distortion of part \cite{ref9}\cite{ref10}\cite{ref11}.} Quality of metal LAM products usually suffer the problem of part porosity, which refers to the pore size, shape, number, and spatial distribution inside the printed parts. \ch{Pore generation is closely correlated with manufacturing processes and environments, including machine setting parameters, Heat-treatments, and testing methods \cite{ref12}\cite{ref13}.} Porosity levels closely influences the printed parts’ mechanical performances, such as tensile/shear strength, hardness, fatigue limit, thermal/electrical conductivity, and corrosion resistance \cite{ref14}\cite{ref15}\cite{ref16}. To assure desired mechanical performances of a part, \ch{it is critical to explore the predictive porosity model for part quality assurance}, based on which part porosity could be accurately predicted and inhibited towards the positive-performance-influence direction.

\ch{Studies have investigated defect generation mechanisms during AM processes \cite{ref9}\cite{ref10}\cite{ref11}\cite{ref12}\cite{ref14}. To ensure optimal machine settings, researchers have experimentally reported the influences of process parameters, such as laser energy density, laser power and speed, scan strategy, layer thickness, and scan spacing on AM-printed parts’ microstructure, mechanical properties, density, and surface quality \cite{ref9} \cite{ref10}\cite{ref17}\cite{ref18}\cite{ref19}. Some researchers focused on optimizing the melt pool behavior with physics mechanics for part quality assurance \cite{ref20}\cite{ref21}\cite{ref22}\cite{ref23}\cite{ref24}. H. Gong analyzed porosity by investigating melt pool sizes and default-setting process parameters, such as laser scanning speed, maximum current, line offset, and focus offset \cite{ref25}. S. Khademzadeh analyzed porosity levels by using machine setting parameters, such as hatch distance, scanning strategies, and powder feeding rate \cite{ref26}\cite{ref27}. The influences of machine setting parameters such as layer thickness, laser scan speed, laser power, and substrate temperature, on pore generation were analyzed. Although these methods can model process – property correlations, a large number of trail-and-error experiments or computationally intense simulations are required, introducing high time and economic cost to limit the experiments scale and further limit the porosity analysis.}

\ch{In recent years, data-driven machine learning (ML) approaches for AM process – property modeling are becoming prevalent \cite{ref28}\cite{ref29}\cite{ref30}\cite{ref31}\cite{ref32}\cite{ref33}\cite{ref34}. The Random Forest was used to model the process – property and process – structure relationships for laser powder bed fusion (LPBF) fabrication of Inconel 708 \cite{ref34}. Bayesian networks effectively modeled process – property relations for knowledge across multiple metals AM printers, and verified through three “industry use” inspired scenarios by multi-properties optimization on laser powder bed fusion Ti-6Al-4V \cite{ref35}. With a Gaussian predictive model, the porosity levels of 17-4 PH stainless steel parts fabricated on a ProX 100TM SLM system were predicted based on influential machine setting parameters, such as laser power and laser scan speed \cite{ref36}. Kamath streamlined the process of parameter optimization and identify parameters using regression tree model algorithms. It was used to create small pillars with 99\% density across a relative larger power range, 150 $\sim$ 400 W \cite{ref37}. A. Garg formulated a model with multi-gene genetic programming for understanding the functional relationship of open porosity of the AM process \cite{ref38}. Although data-driven ML approaches are well known for their extraordinary ability to learn the complex process – property correlations, their models are either machine-setting dependent or ignore the underlying physics analysis of pore generation. When the machine, components and/or ranges of machine settings are changed, this already built prediction models are often failed. The fundamental reason for this limitation is that machine-dependent setting parameters are indirect for pore generation mechanism. The direct reasons were heat and thermal pressure histories of laser-material interaction times during AM process \cite{ref39}. A brute-force exploration of machine setting parameters only brings very large design space but limit the envision of governing physics mechanics of pore generation. Without exploring the knowledge underlying physics phenomenon of pore generation, it is difficult to investigate the microstructure and macroscopic mechanical properties variations brought by various types of machine setting parameters, give rise to the limitation of general implementation of machine setting models for porosity analysis.}

\ch{Nowadays, people are paying more attention on physics-informed data-driven modelling \cite{ref40}\cite{ref41}\cite{ref42}\cite{ref43}\cite{ref44}\cite{ref45}. \\One way is providing training datasets from physics-based simulations to ML models. For example, Yan \cite{ref43} concepted the idea of physics-informed data-driven modeling framework to derive process-structure-property relationships for AM. Wang \cite{ref42} recently implemented this framework for comprehensive AM uncertainty quantification, in which a large number of multiscale physics-based simulations such as Finite-element based thermal model and phase-field model, to generate training data for construction of computationally inexpensive ML models. The other way in material informatics community is to engineer physical meaningful features as input for ML model, to improve the model performance with limited amounts of experimental datasets. For example, the physics-informed composition features were the key to data-driven models to predict the glass-forming ability of metallic glasses \cite{ref45}, band gap energies of materials \cite{ref46}, formation enthalpies of semiconductors \cite{ref47}, and critical temperatures of superconductors \cite{ref48}. In addition to physics-informed composition features, Liu \cite{ref40} recently engineered physics-informed heat-treatments process features for composition – process – property relationships of shape memory alloys design.”}

\begin{figure}[ht]
\centering
\includegraphics[width=0.9 \linewidth ]{ 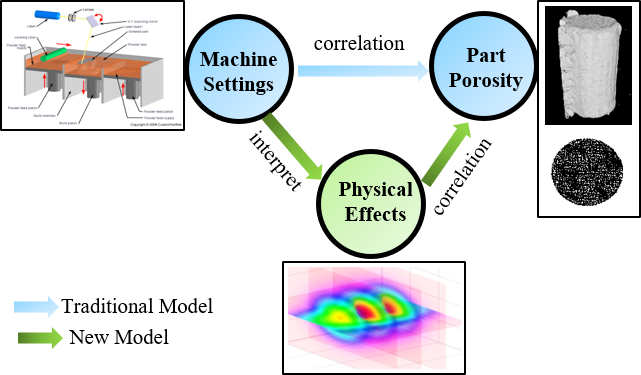}
\caption{\ch{A novel physics-informed data-driven model. Instead of usingmachine setting parameters (in blue arrow), such as laser speed, laser power, part pose, part location, the novel model uses physical effects of machine setting parameters(in green arrow), such as energy density distribution and pressure distribution, to  model process – porosity correlations.}}
\label{fig:1}
\end{figure}

\ch{To incorporate the physics effects into the data-driven model and provide in-depth insights on the porosity generation analysis, in this paper, a physics-informed data-driven porosity prediction model is developed for AM process.} To validate the developed methodology, laser-related physics effects are chosen as an example for porosity analysis. Instead of directly using machine-related setting parameters, this model can analyze pore generation by using physical effects during short laser-material interaction times, which explore essential pore generation underlying mechanisms from the laser energy and radiation pressure perspectives, as shown in Fig. \ref{fig:1}. The intrinsic, influential physical effects are automatically identified in the correlation modeling process. Various machine setting parameters are interpreted into their physical effects first, and then a physics-informed data-driven model is built to learn the correlation between these physical effects and porosity sizes. By identifying influential physical effects and their influential ranges, the pore generation mechanism is analyzed. 

\ch{Since the physics-informed effects during LAM laser-materials interaction process are machine independent, the developed physics-informed data-driven model is flexible for deployment and  has the potential to be scalable on different types of laser-based AM systems. Our contributions could be summarized as two-fold.
\begin{enumerate}
\item A generalizable physics-informed porosity prediction m-odel (PIM) is developed for porosity estimation and prediction. Instead of directly using machine-setting parameters for modeling, the physics-informed model uses fundamental physical effects underlying machine settings to build a generalizable prediction model. \ch{The physics-informed model has comparable performance with machine setting model in our study of 3D printed part porosity analysis on Concept Laser M2 machine;}
\item A physical effect perspective is introduced into machine-learning-based porosity prediction, which help classify the processing space and identify subregions where pores formed. This work shows that physics-informed ML mod-el approaches extract general physics effects in LAM process to improve laser-based AM modeling capabilities.
\end{enumerate}}

\section{Physical effect calculation}
\subsection{Metal SLM experiment setup}
\begin{figure}[ht]
    \centering 
    \subfigure[Laser and part setting]
    {
        \begin{minipage}[t]{0.48\linewidth}
            \centering  
            \includegraphics[width=1\linewidth]{ 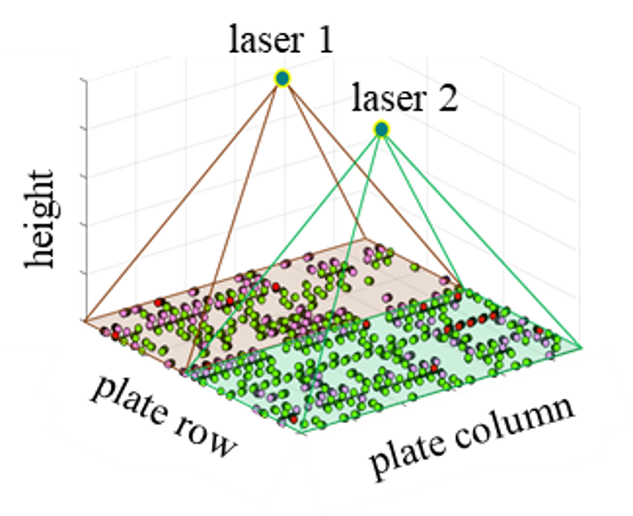} 
        \end{minipage}%
    }
    \subfigure[Building plate]
    {
        \begin{minipage}[t]{0.48\linewidth}
            \centering 
            \includegraphics[width=1\linewidth]{ 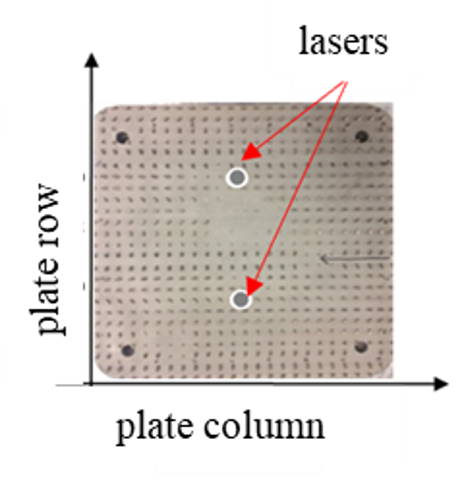}
        \end{minipage}
    }
    
    \subfigure[Parts with poses]
    {
        \begin{minipage}[t]{0.48\linewidth}
            \centering  
            \includegraphics[width=1\linewidth]{ 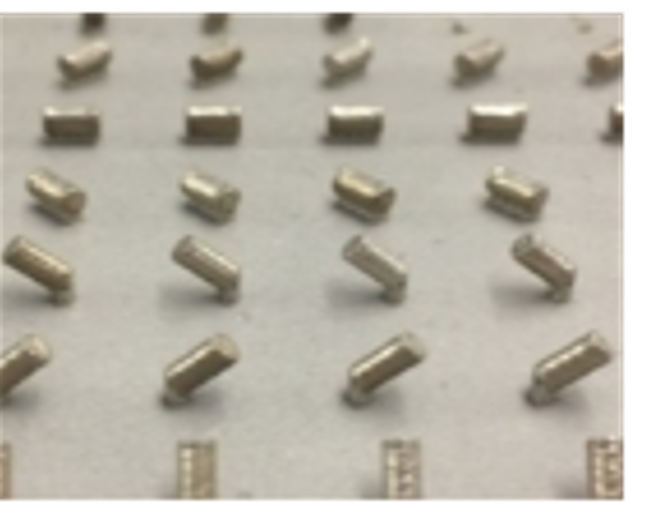} 
        \end{minipage}%
    }
    \subfigure[Part modeling]
    {
        \begin{minipage}[t]{0.48\linewidth}
            \centering 
            \includegraphics[width=1\linewidth]{ 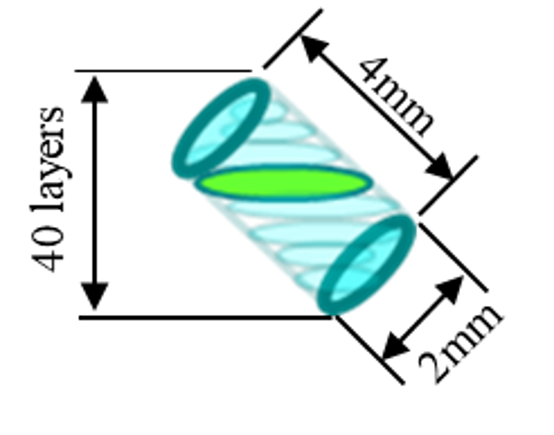}
        \end{minipage}
    }
    \caption{Schematic diagram of a four-level tripod-type atomic system driven by three coherent laser fields.} 
    \label{fig:2}
\end{figure}

The metal SLM samples were produced on a Concept Laser M2 Dual-Laser Cusing machine. This system has a 27.7 x 27.7 x 30 $cm^3$ build envelope. Inconel 718 powder was spread from the supply chamber over the build area using a rubber wiper blade. \ch{As with other metal AM techniques, SLM uses an inert gas environment to mitigate oxide and inclusion formation. The active flow of Ar was switched on when the oxygen level is greater than 0.6\%.} A 200 $W$ fiber laser is directed by a gimbal set 482.6 $mm$ from the build surface. The selected laser parameters were as follows: serpentine track pattern at 160 $W$ laser power, 800 $mm/s$ laser speed and a 50 micron spot size (FWHM). Layer thickness is set as 50 microns. The dual-laser Cusing uses two lasers to divide the exposure surface in half and assign to each laser one half of the plate area. \ch{The build volume was maintained in Argon atmosphere with temperature in the range of 21 - 27 $^{\circ}$C and relative humidity in range of 6 - 18\%. To investigate the porosity distribution in different locations of an AM part, small cylinders with diameter 2 $mm$ and height 4 $mm$ were build across the print plate.} The build geometries are shown in Fig. \ref{fig:2}(a). A plate is visualized as that in Fig. \ref{fig:2}(b), in which two dots denote laser’s vertical projections. The build plate has dimensions of 247 $mm$ x 247 $mm$. 9 unique orientations of samples were built on the plate. Every row on the plate is a unique orientation and each orientation is involved in the whole plate, shown in Fig. \ref{fig:2}(c). In total, 605 samples were built on the plate. We then selected 549 parts for porosity analysis. To conduct layer-wise part porosity analysis, a part is simplified as 40 horizontal layers, evenly distributed along the vertical direction with respect to a part’s original setting pose, shown in Fig. \ref{fig:2}(d).

\begin{figure}[ht]
\centering
\includegraphics[width=0.9\linewidth ]{ 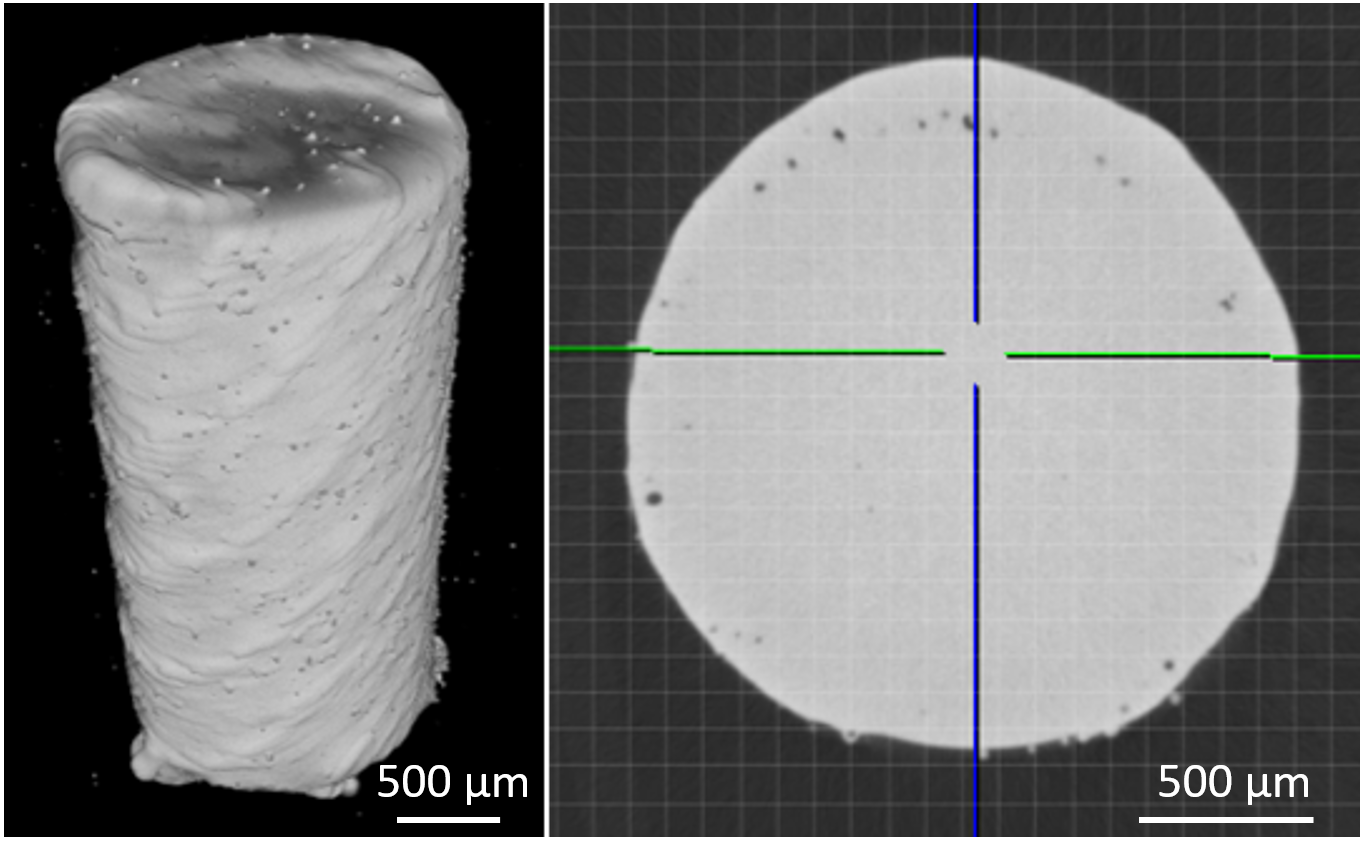}
\caption{\ch{A x-ray tomography of cylinder shape sample with diameter 2 $mm$ and height 4 $mm$. The topographical view (left) renders the sample’s as-built surface, a cross sectional view (right) reveals subsurface porosity.}}
\label{fig:3}
\end{figure}

\begin{figure}[ht]
\centering
\includegraphics[width=0.9 \linewidth ]{ 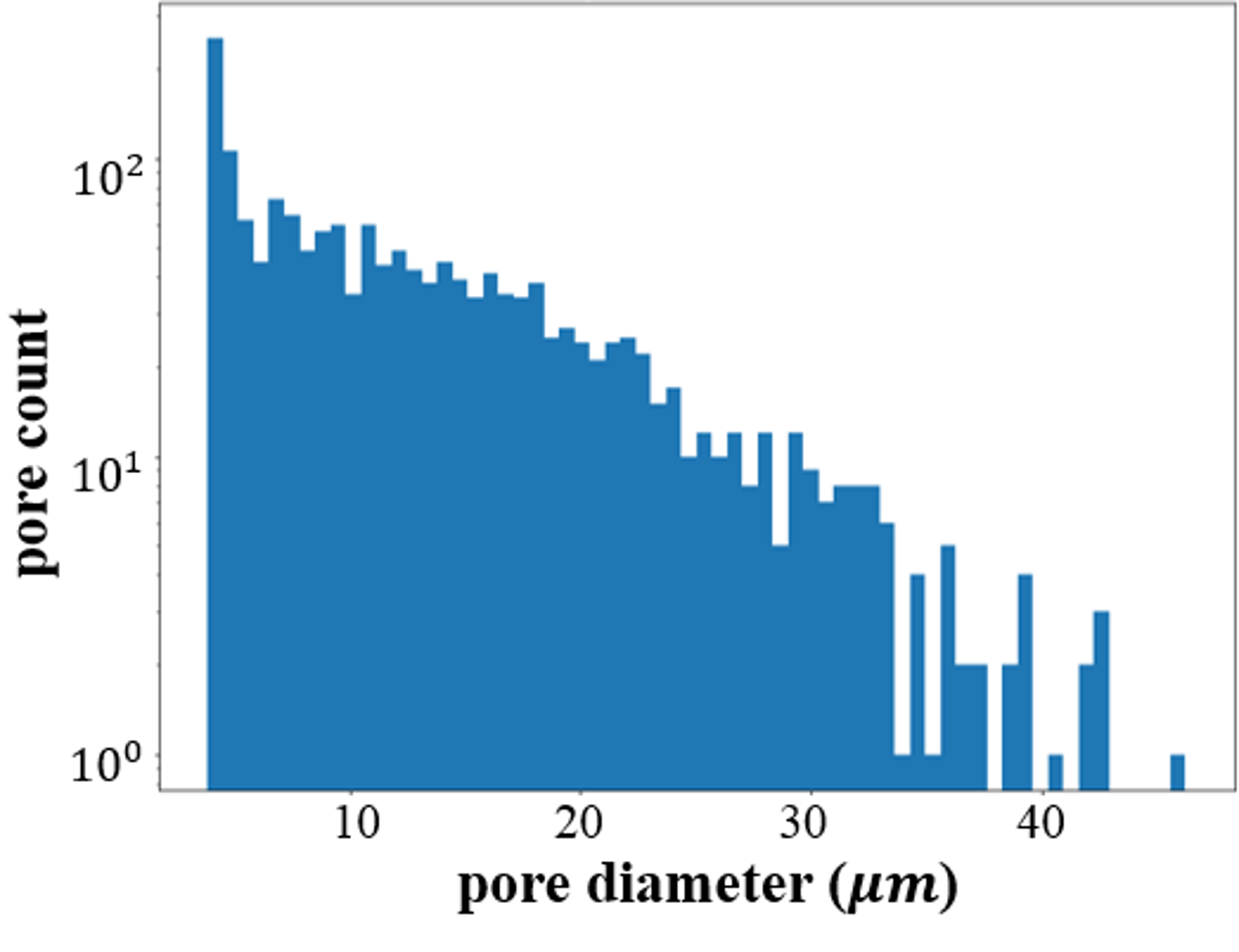}
\caption{A histogram representation of effective pore diameters (in microns) for a single sample. As can be seen, porosity visible using X-ray micro-CT is on the order of 1’s - 10’s of microns in diameter and up}
\label{fig:4}
\end{figure}

To inspect the porosity networks in the parts, each was initially scanned using micro X-ray computed tomography (XCT). Using the raw tomography data as shown in Fig. \ref{fig:3}, an in-house batch analysis routine, Tomography Reconstruction Analysis and Characterization Routines (TRACR) \cite{ref49}, were used to identify and analyze internal porosity using a series of image processing steps and statistical tools. Response parameters such as maximum pore diameter, mean pore diameter, median pore diameter (Fig. \ref{fig:4}), and median pore spacing, were derived using TRACR.

\subsection{Laser physical effects calculation}
\begin{figure*}[t]
\centering
\begin{minipage}[t]{0.3\linewidth}
\centering
\includegraphics[width=0.9\linewidth]{ 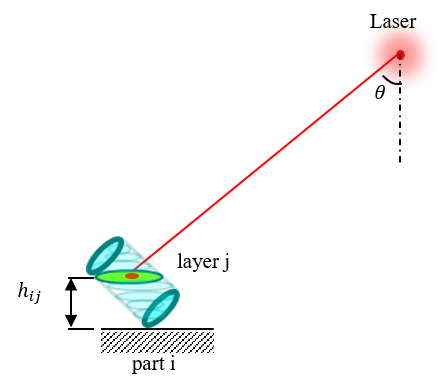}
\caption{World Map}
\label{fig:5}
\end{minipage}
\begin{minipage}[t]{0.3\linewidth}
\centering
\includegraphics[width=0.9\linewidth]{ 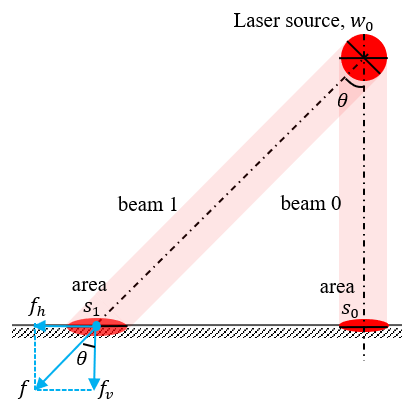}
\caption{Concrete and Constructions}
\label{fig:6}
\end{minipage}
\begin{minipage}[t]{0.3\linewidth}
\centering
\includegraphics[width=0.9\linewidth]{ 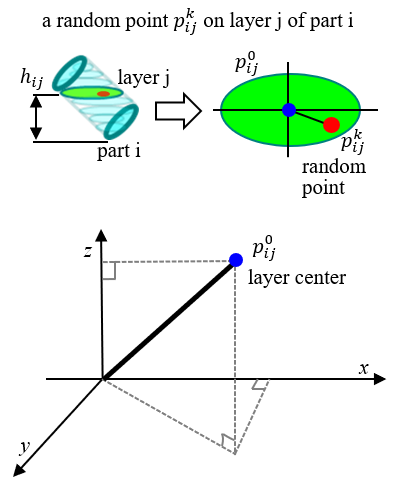}
\caption{Concrete and Constructions}
\label{fig:7}
\end{minipage}
\end{figure*}

Even though many physical effects, such as powder energy absorption and material/air heat transfer characteristics, influence pore generation, for simplification we only consider laser radiation, pressure intensity, and laser energy density during manufacturing processes in this paper. Spatial relations between a part and a laser are shown in Fig. \ref{fig:5}, where a laser is melting a random layer $j$ with height $h_{ij}$ of part $i$. $\theta$ is the laser scanning angle, defined by the angle between the laser projection direction and the vertical direction.
First, energy density on a random point of the powder bed is analyzed. Laser source area is $s_0$, which is also the projection area of the laser beam in the vertical direction ($\theta=0$), shown in Fig. \ref{fig:6}. When a laser is scanning a random point, the laser beam projection area is simplified as an ellipse, calculated by Eq. 1, shown in Figs. \ref{fig:6} and \ref{fig:7}. During laser scanning, laser power $w_0$ is constant. Although laser energy distribution is not even on a projection area, we simplify it as such to simplify the calculation. Therefore, the average energy density in a random point on the powder bed is calculated by Eq. 2, where $e_0$ is energy density in the vertical direction, and $\theta$ is the laser angle.

Then, energy density $e_{p_{ij}^k}$ on a specific point $p_{ij}^k$ in layer j of part i is calculated by Eq. 3, where $\theta_{p_{ij}^k}$ is the laser angle when point $p_{ij}^k$ is scanned. Based on the layer center $p_{ij}^0$, position of a random point $p_{ij}^k$ is calculated by referring the position of layer center $p_{ij}^0$.
 
 
\noindent Projection area
\begin{equation}
s_1=s_0/cos\theta
\end{equation}    
Power intensity	
\begin{equation}
e_1=\frac{w_0}{s_1}=w_0 cos\theta/s_0=e_0 cos\theta
\end{equation}
Energy density	
\begin{equation}
e_{p_{ij}^k}=e_0 cos\theta_{p_{ij}^k}=w_0 cos\theta_{p_{ij}^k}/s_0
\end{equation}

\begin{figure}[ht]
\centering
\includegraphics[width=0.6 \linewidth ]{ 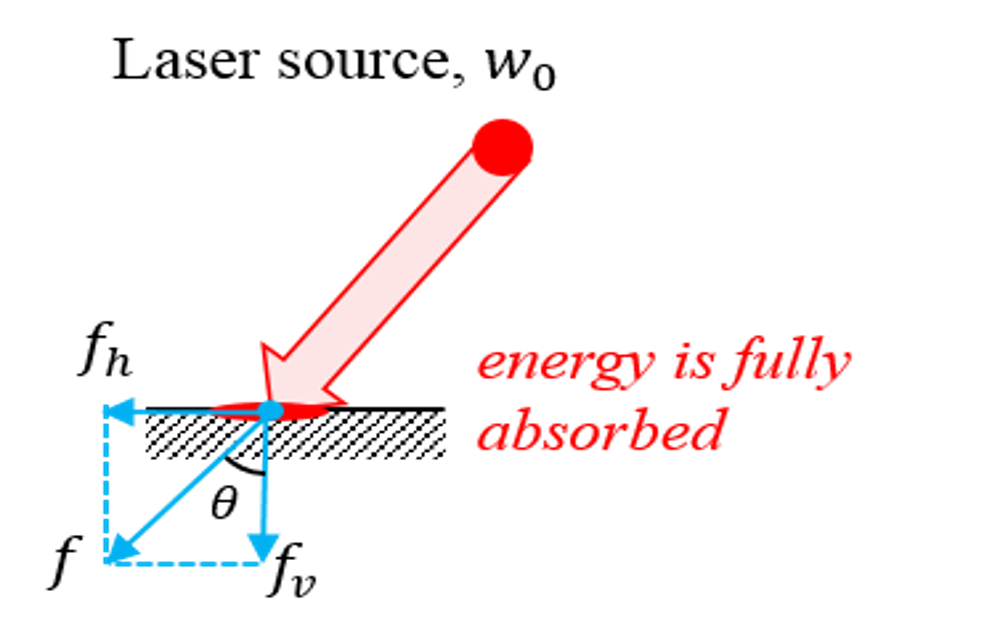}
\caption{Laser radiation pressure calculation. When a laser angle is $\theta$, a radiation pressure is generated on the layer surface. To investigate pressure influence on keyhole phenomenon, absolute pressure is decomposed into horizontal pressure and vertical pressure.}
\label{fig:8}
\end{figure}

A laser beam exerts radiation force on the powder bed surface. To simplify our model, we assume all the energy and the momentum of the laser beam is fully transferred to the melt pool. That means, all the momentum of the laser beam is fully transferred to the radiation force, shown in Fig. \ref{fig:8}. In a practical situation, the absorbed energy is a fixed proportion of a laser beam, and all the data are normalized for training our model. Therefore, this simplification does not significantly influence physical-effect-based porosity prediction. As Eqs.  (4-11) expressed, each photon has energy $E_0$ (Eq. 4) with momentum $p_0$ (Eq. 5) \cite{ref15}\cite{ref16}, and where $c$ is the speed of light in $a$ vacuum, $h$ is Planck constant and $\lambda$ is the laser light’s wavelength. For each second, the total number $N$ of photons projected by a laser beam is calculated by laser power $w_0$(160 $J/s$) divided by a photon’ power $E_0$, shown in Eq. 6. The total radiation force, $F_N$, exerted by a laser beam is calculated by the derivative value of all the $N$ photons’ momentum $P$ given time $t$, shown in Eq. 7. The final radiation pressure $f$ exerted by a laser beam is calculated by the total force, $F_N$, divided by the laser beam’s projection area, $s_1$, shown in Eq. 8. Based on the known laser angle $\theta$, pressure components along vertical and horizontal directions have been calculated by Eqs. 9 and 10, respectively.

\noindent Energy of one photon 
\begin{equation}
E_0=hc/\lambda
\end{equation}
Photon momentum
\begin{equation}
p_0=E_0/c=h/\lambda
\end{equation}
Total photon number $n$ ejected in each second
\begin{equation}
N=w_0/E_0=w_0 \lambda/hc
\end{equation}
$N$ photons have momentum $P$, total radiation force exerted by $N$ photons is $F_N$
\begin{equation}
\begin{split}
F_N&=dP/dt\\
&=d(Np_{0}t)/dt\\
&=Np_0=\frac{w_0 \lambda}{hc}\frac{h}{\lambda}=w_0/c
\end{split}
\end{equation}
Radiation pressure $f$ at point $p_{ij}^k$
\begin{equation}
f=\frac{F_N}{s_1}=\frac{w_0}{cs_0}cos\theta_{p_{ij}^k}
\end{equation}
, where $\theta_{p_{ij}^k}$  is the incident angle.

\noindent Vertical pressure
\begin{equation}
f_v=\frac{w_0}{cs_0}cos^2 \theta_{p_{ij}^k}
\end{equation}
Horizontal pressure 
\begin{equation}
f_h=\frac{w_0}{cs_0}cos\theta_{p_{ij}^k}sin\theta_{p_{ij}^k}
\end{equation}

\section{Physics-informed porosity prediction model}
\subsection{Porosity prediction}
Based on the laser physical effects, a physics-informed data-driven model for analyzing part porosity levels was designed. To evaluate the porosity prediction performances of the model and investigate the appropriate manner of using physical effects in porosity prediction, two baseline models were involved: machine-setting model (‘setting’ for short) and \\machine-setting-physical-effect combined model (‘combined’ for short). Inputs of the physics-informed are physical effects related to energy density and pressure on a part. Inputs of the setting model are the spatial and geometry parameters related to the plate, part, laser and layers. Inputs of the combined model are the combinations of machine setting parameters and physical effect features. For all the three types of models {setting model, combined model, physics-informed model}, the output is one of the four types of porosity levels {maximum pore diameter, mean pore diameter, median pore diameter, median pore spacing}. Maximum pore size influences part quality. Median pore spacing is decided by pore distribution. Diameter based calculations (maximum, mean, median) were made using a spherical approximation of the pore volumes. Pore spacing was calculated using a nearest neighbor search of centroid-centroid positions. To support each model and find the optimal learning algorithm, six classic regression algorithms, which were previously used in material property prediction and achieved good performances, are adopted in the porosity prediction \cite{ref52}. Six regression algorithms were tested to select the most suitable algorithm for the optimal performance. They are linear regression (Linear), Gaussian Process regression (GPR), support vector regression with linear kernel (svrR), support vector regression with Gaussian kernel (svrG), support vector regression with radial basis function kernel (svrRBF), and support vector regression with polynomial kernel (svrP) \cite{ref53} \cite{ref54}. Modeling process is detailed below.

\textit{Linear}. Linear regression has the form shown in Eq. 11. $X\in R_{n\times m}$ is input features; $Y\in R_{1\times m}$ is a specific porosity level. $W\in R_{n\times 1}$ are coefficients of inputs $X$ in the linear regression algorithm.
\begin{equation}
    Y=W^{T}X+\epsilon
\end{equation}

\textit{GPR}. GPR is with the form in Eq. 12.
\begin{equation}
    Y=f(X)+N(0,\sigma_{n}^2)
\end{equation}
where $f(X)$ is covariance function \\$k(X, X^{'})= \sigma_{f}^{2}exp[\frac{-(X-X^{'})^{2}}{2l^{2}}]+\sigma_{n}^{2}\delta(X,X^{'})$. $f$ is an infinite Gaussian distribution denoted by the mean 0 and variance $\sigma_n^2$ of input features $X$. The covariance $\delta$ is calculated by the distance between features in $X$.

\textit{SVR}. Training data is $\{(x_1,y_1), …, (x_l,y_l)\}\in X$. $x_l$ is the input features of part $i$, and $y_l$ is the porosity level of part $i$. By using SVR, the goal is to search a function $f(x)$, which can predict porosity levels for all the parts with a desired prediction error $\epsilon$. A general format of SVR is shown in Eq. 13. $k$ is the kernel function used to fit input features in a specific manner. $w$ is the coefficient of features in the kernel-function-transformed feature space, and $b$ is the interception of function $f(x)$. $x$ and $x^{'}$ denote two different input features within a feature type. Cost function is defined by Eq. 14. With the cost function, the total accumulated errors are reduced by searching the optimal weight parameters $w$ and optimal slacks $\xi_i$ and $\xi_{i}^{'}$. With the prediction error constraints $\epsilon$, and error slacks $\xi_i$, $\xi_{i}^{'}$, the value differences between each correctly predicted porosity level and its actual porosity level $y_l$ are required to be within the error limit $\epsilon$, shown in the constraint Eqs. 15 and 16. The regression process is a process of solving the quadratic programming (QP) problem with the cost function shown in equation 14 and constraints in Eqs. 15 and 16. To explore accurate correlations among input features and part porosity levels with the consideration of computational cost reduction, different linear and nonlinear kernels are used to explore patterns in input features. Eq. 17 is a linear kernel, which explores input feature correlations by dot product. Wqs. 18, 19, and 20 are Gaussian kernel, radial basis function kernel, and polynomial kernel, respectively, which explore input feature correlations by a nonlinear function. With these nonlinear functions, raw input features such as machine settings and laser physical effects can be transformed and projected into a higher dimension, where the hidden correlations among features are easier to be explored for improving accuracy of correlation modeling.

\begin{align}
    &f(x,w)=\sum_{j=1}^{m}w_{i}k(x,x^{'})+b\\
    &\textbf{min}\frac{1}{2}||w||^{2}+C\sum_{i=1}^{n}(\xi_i+\xi_i^{'})\\
    &\textbf{s.t.}\forall i:|y_{i}-w^{T}k(x,x^{'})-b|\leq \epsilon+\xi_{i}\\
    &|y_{i}-w^{T}k(x,x^{'})-b|\leq \epsilon+\xi_{i}^{'}
\end{align}
\noindent Linear kernel
\begin{equation}
    k(x,x^{'})=x\cdot x^{'}
\end{equation}
Gaussian kernel
\begin{equation}
    k(x,x^{'})=exp(-||x-x^{'}||^{2})
\end{equation}
RBF kernel
\begin{equation}
    k(x,x^{'})=exp(-||x-x^{'}||^{2}/2\sigma^{2})
\end{equation}
Polynomial kernel
\begin{equation}
    k(x,x^{'})=(1+xx^{'})^{p}
\end{equation}

\subsection{Quality-Assured porosity analysis}
\ch{To account for different pore generation mechanisms that may be present during the build process, the porosity of each part is heuristically subdivided into “pass” (pores with maximum size up to diameter of 97 $\mu m$), “flag” (97 $\mu m$ $\leq$ maximum diameter $<$ 220 $\mu m$) and “fail” (pore maximum diameter $\geq$ 220 $\mu m$). As shown in Fig. 4, for a single part’s pore size distribution, most of pores size is less than 30 $\sim$ 40 $\mu m$. The pores of part with size larger than 30 $\sim$ 40 $\mu m$ only take a little fraction of total pores inside part. Some parts with largest pore size few of hundred micros, this may result from sample removing and preparation processes, the interface of part and plate substrate exists a lot of large pore size.  In order to model maximum pore size prediction, we define the pore with size larger than $\sim$10\% of the minimum geometric parameter of the part can be treated as fail (i.e., in this paper, a 220 $\mu m$ pore would be a fail considering the 2 $mm$ diameter of the test pins). A pore less than 5\% of the minimum geometric parameter of the part is a pass (i.e., 97 $\mu m$ in this paper).} A porosity analysis model is created to predict porosity value and investigate physical influence for each quality level, instead of analyzing the physical influence for all the samples in the full value range. According to our previous experiences in part porosity analysis, part pore size distribution is usually sparse, indicating that the value range is wide while porosity values are concentrated in some regions. From a modeling perspective, sparse porosity distribution brings noise into the model training process, decreasing model prediction accuracy. By categorizing data into different concentrated data clusters for training each model individually, the model will be more focused with improved performances.
To quantify model error, we adopted the N-fold cross validation method \cite{ref55}. The finalized error after cross-validation is defined by Eq. 21. $y$ is an actual porosity value, and $y^{'}$ is a predicted porosity value. $g(y,y^{'})$ is a generalized error metric, which could be absolute error, $|y – y^{'}|$; standard error, $|y-y^{'}|/\sigma$ ; or percentage error, $|y-y^{'}|/y$. The data set was randomly divided into n number of folders. At each time $n-1$ folders were selected to train the model, the remaining 1 folder was used to test the learned porosity prediction model. The training-testing process was iteratively conducted by selecting different training folders and testing folders until every folder participated in both testing and training. After n-fold cross validation, the average porosity prediction error $\epsilon_c$ is calculated.

\begin{equation}
    \epsilon_c=\frac{1}{n}\sum_{i=1}^{n}g(y,y^{'})
\end{equation}

\section{Results and discussion}
\subsection{Porosity distribution visualization}

\begin{figure}[ht]
\centering
\includegraphics[width=0.9 \linewidth ]{ 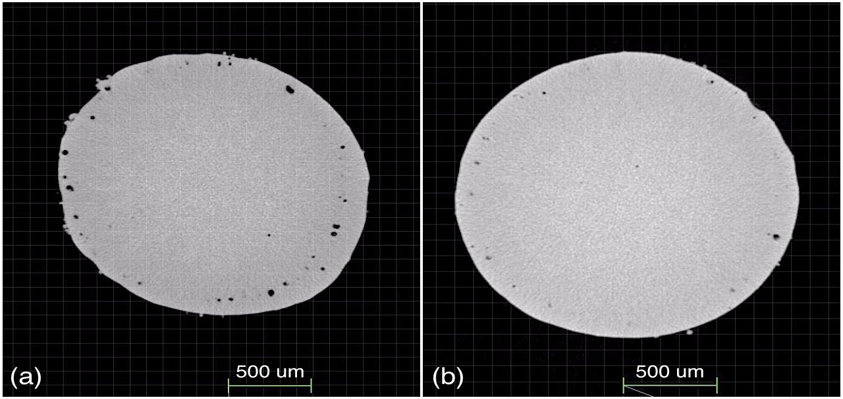}
\caption{\ch{A tomography frame showing internal porosity of sample with different  porosity size levels.}}
\label{fig:9}
\end{figure}

\begin{figure*}[ht]
\centering
\includegraphics[width=0.7 \linewidth ]{ 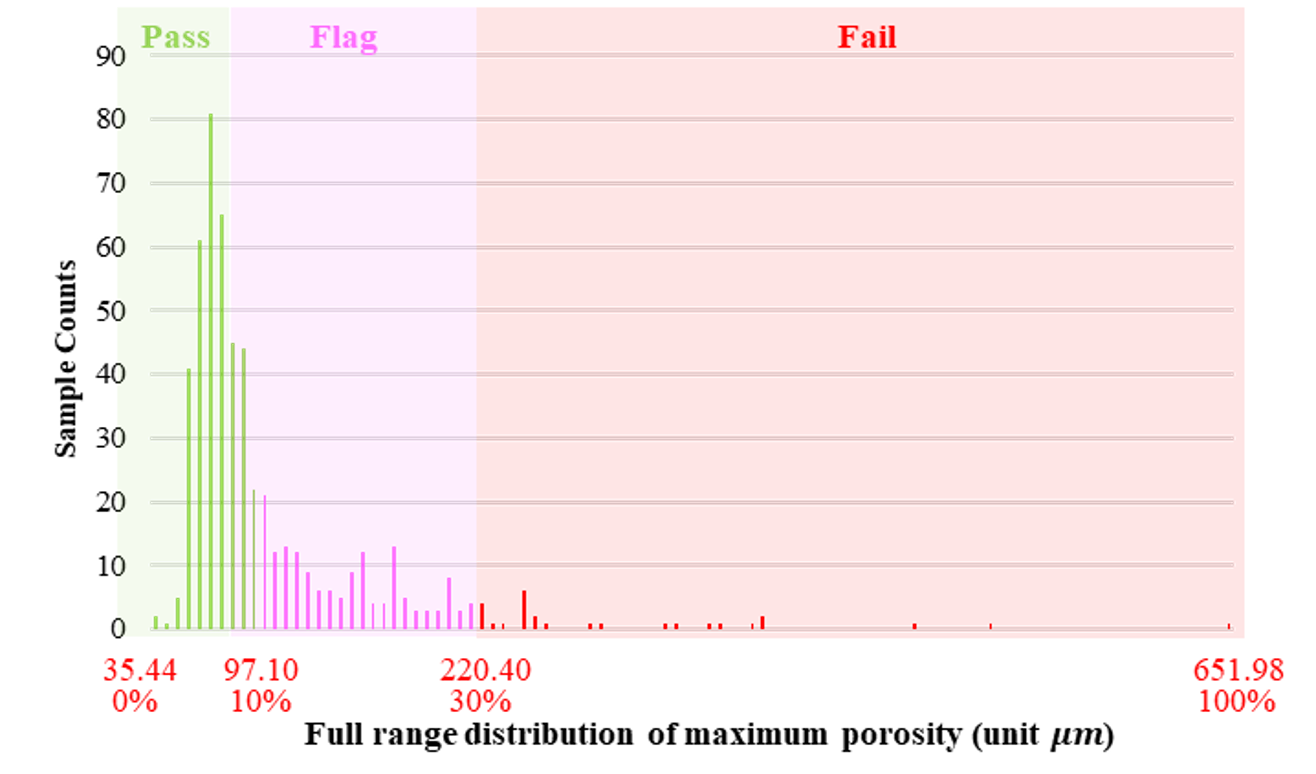}
\caption{\ch{The distribution of maximum pore diameter of each part on the print plates, in total 549 parts are studied among 605 printed parts. One single part’s pore size distribution is illustrated in Fig. 4. Most of the parts are with maximum pore size less than 100 $\mu m$. According to domain experts’ experience, parts with a maximum pore size below 97.10 $\mu m$ are considered to be “pass” parts, which means the quality exceeds the product expectation; parts with a maximum pore size ranging in [97.10 $\mu m$, 220.40 $\mu m$] are considered to be “flag” parts, which means the quality is in the margin area; parts with a maximum pore size larger than 220.40 $\mu m$ are considered to be “failed” parts, which means the part is failed. Percentage value is to show the relative size of a pore comparing with the smallest maximum pore size (defined as 0\%) and the largest maximum pore size (defined as 100\%).}}
\label{fig:10}
\end{figure*}

\ch{X-Ray tomography are used for identifying part porosity size values (Fig. \ref{fig:9}) and four porosity metrics - maximum pore diameter, mean pore diameter, median pore diameter and median pore spacing. Each black dot inside the part’s cross section denotes a pore. Dot sizes are proportional to porosity size levels. }For the maximum pore diameter, most parts have a maximum pore diameter of 35.44 $\mu m$ $\sim$ 189.58 $\mu m$ and a mean pore diameter of 16.17 $\mu m$ $\sim$ 20.60 $\mu m$. The distribution of median pore diameters is similar to that of the mean pore diameters, which were 15.13 $\mu m$ $\sim$ 19.90 $\mu m$. \ch{For the median pore spacing, adjacent pores are separated being about 54.87 $\mu m$ $\sim$ 63.78 $\mu m$ . Comparing all four of these porosity types, either larger pores or smaller pores were generated across the build plate. Also, relatively higher levels of porosity were more frequently appeared in the plates’ corners and borders. For details, please see the attachment file, porosity visualization.}

\subsection{Physical effect calculation}
To describe physical effect distribution on a whole part, four quantitative indicators were adopted to summarize physical effect distributions on a part’s 40 layers: average physical effect (AVE), standard deviation of physical effect (SDEV), minimum physical effect (MIN) and maximum physical effect (MAX).
Given the height of the part is 4 $mm$, we slide the part into 40 layers to estimate the internal physical effect distribution. AVE was calculated by the overall arithmetic mean of the physical effect values on all 40 layers. SDEV was calculated by the deviation of layer-wise physical effect values. MIN was calculated by the arithmetic mean of the minimal 10\% physical effect values of all the 40 layers. MAX was calculated by the arithmetic mean of the maximum 10\% physical effect values of all the 40 layers.

Results for energy density distribution are presented as a prototypical demonstration of the data analysis processes. The visualization of physical effect results is shown in the appendix. Energy density distribution and absolute radiation pressure distribution are consistent because their values are proportional, shown in Eqs. 3 and 8. For parts closer to a laser, the energy and force concentrations (revealed by AVE, MIN, MAX values. (a), (c), (d)) are higher, while the energy and force variances (revealed by SDEV values) are lower. For vertical pressure (in the appendix), distribution patterns are consistent with absolute pressure distributions, while the vertical pressure values are less. For horizontal pressure, its distributions on parts are in inverse proportion to vertical pressure distributions. Horizontal force concentrates on the far corners of each half of a build plate. At the center area that is close to a laser, horizontal force distributions are with high variance. As the laser working power is consistent and laser positions are fixed, laser energy density decreases by the increase of its projecting angle, which increases the laser beam’s projection area. This consistent distribution of laser energy and porosity shows the calculations of these physical effects are reasonable.

\subsection{Porosity prediction performances}

\begin{figure*}[ht]
\centering
\includegraphics[width=1 \linewidth ]{ 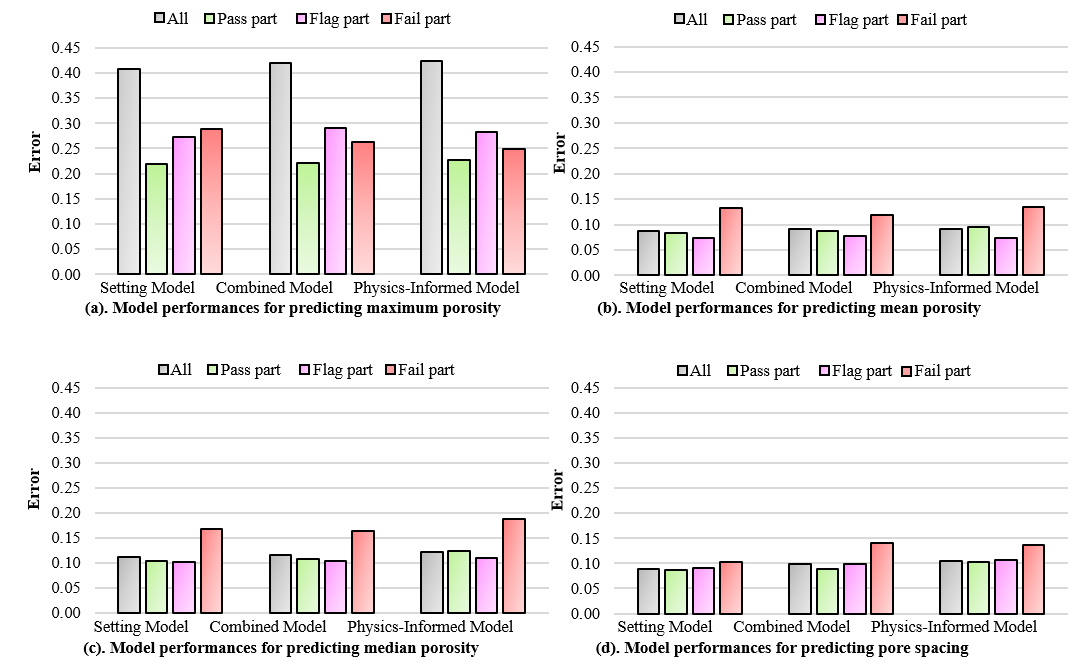}
\caption{Averaged performances for the three types of porosity prediction models. Setting models have the best performances compared with the physics-informed model and combined model, while the performance differences among the three models are slight within error 3\%. Models’ performances are consistent that for maximum pore size the performance is at about 27\% while for other measures the performance is at about 12\%. After categorizing parts into three quality levels, model performances for all the four types of porosity measures are improved generally. For the maximum pore size, mode performances gain the largest improvement (about 20\% error reduction comparing with models using Pass-quality parts). }
\label{fig:11}
\end{figure*}

\begin{figure}[ht]
\centering
\includegraphics[width=1 \linewidth ]{ 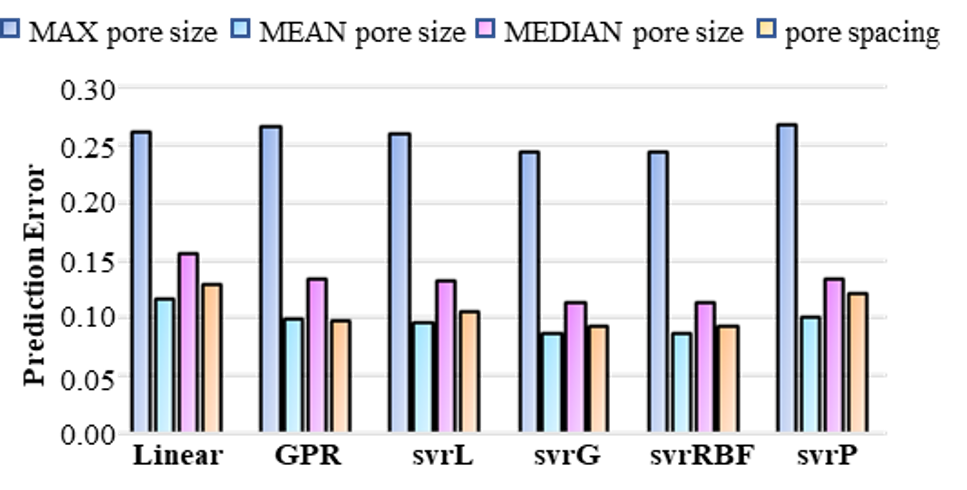}
\caption{Average performances for the involved six algorithms in predicting the four types of porosities. \ch{The algorithms svrG and svrRBF show best performance during the six involved algorithms with an averaged error 13\%.}}
\label{fig:12}
\end{figure}

According to the quality control experience in practical manufacturing, three quality levels of the parts are categorized heuristically into three categories \ch{as introduced in Section 3.2, and shown in Fig. \ref{fig:10}. }

\begin{figure*}[h]
\centering
\includegraphics[width=0.8 \linewidth ]{ 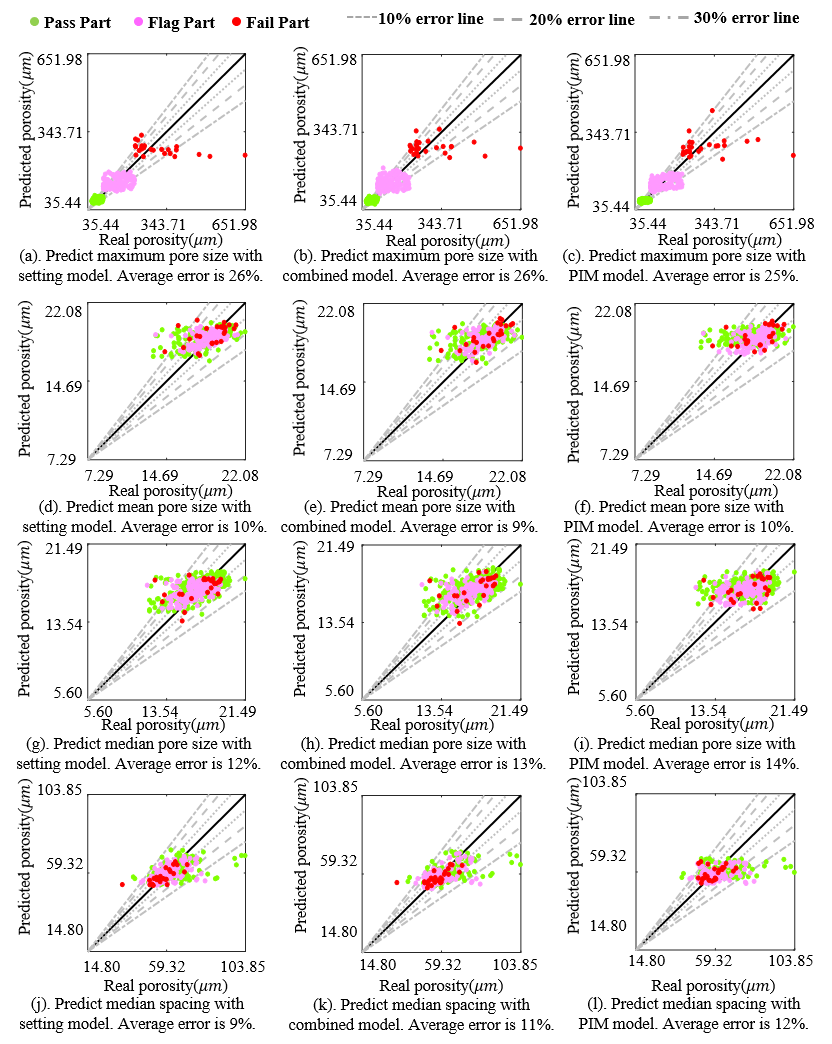}
\caption{Investigating influence of data quality on porosity prediction performances by exploring correlations between predicted porosity and real porosity (PIM: physics-informed porosity analysis model). Based on Figures (a)$\sim$(c), porosity prediction performance is mainly influenced by larger pore sizes ($>$97.10 $\mu m$); based on Figures (d)$\sim$(i), porosity prediction performance is mainly influenced by relatively smaller pore sizes (15.00 $\mu m$); based on Figures (j)$\sim$(l), porosity prediction performance is mainly influenced by both large and small pore sizes ($<$54.87 $\mu m$ and $>$63.78 $\mu m$). Basically, these influential pores are triggered by more influential factors such as blade speed, part inspection methods, and uncertain system errors, instead of merely laser-related physical effects. Analysis of the cause of performance deviation, instead of using black-box machine learning model, helps further understanding of pore generation mechanism and give suggestions for future model improvements.}
\label{fig:13}
\vspace{-1em}
\end{figure*}

Porosity prediction accuracy for the three models are shown in Fig. \ref{fig:11}. The accuracy is denoted by the percentage error of a predicted porosity value given a real porosity value. A lower error means a predicted porosity level is relatively closer to the real porosity level. The setting model has the best performance with the lowest prediction errors of 22\% for maximum porosity, 7\% for mean porosity, 10\% for median porosity, and 9\% for pore spacing. Compared with the setting model, the physics-informed model has a similar performance with only 3\% accuracy derivation. Given the physics-informed model does not rely on machine settings while it has similar performances with the setting model, it holds great potential in being adopted in quality studies with various different machines and/or various machine settings. The combined model has slightly better performance (1\%$\sim$2\% performance improvement) than the physics-informed model, while slightly worse performance (1\%$\sim$2\% performance derivation). The median performance of the combined model shows an effective way of using physical effects-combining physical effects with available machine setting parameters could improve porosity prediction performance. This gives guidance on future implementation of physics-informed porosity analysis models. Compared with the model using all the samples, performances of models using categorized samples are largely improved with about 1\%$\sim$19\% error reduction; meanwhile, performance for “pass” samples achieves the best performance with error between 8\%$\sim$23\% (1\%$\sim$7\% error reduction as compared with the model using “flag” or “fail” part).

\begin{figure*}[ht]
\centering
\includegraphics[width=0.95 \linewidth ]{ 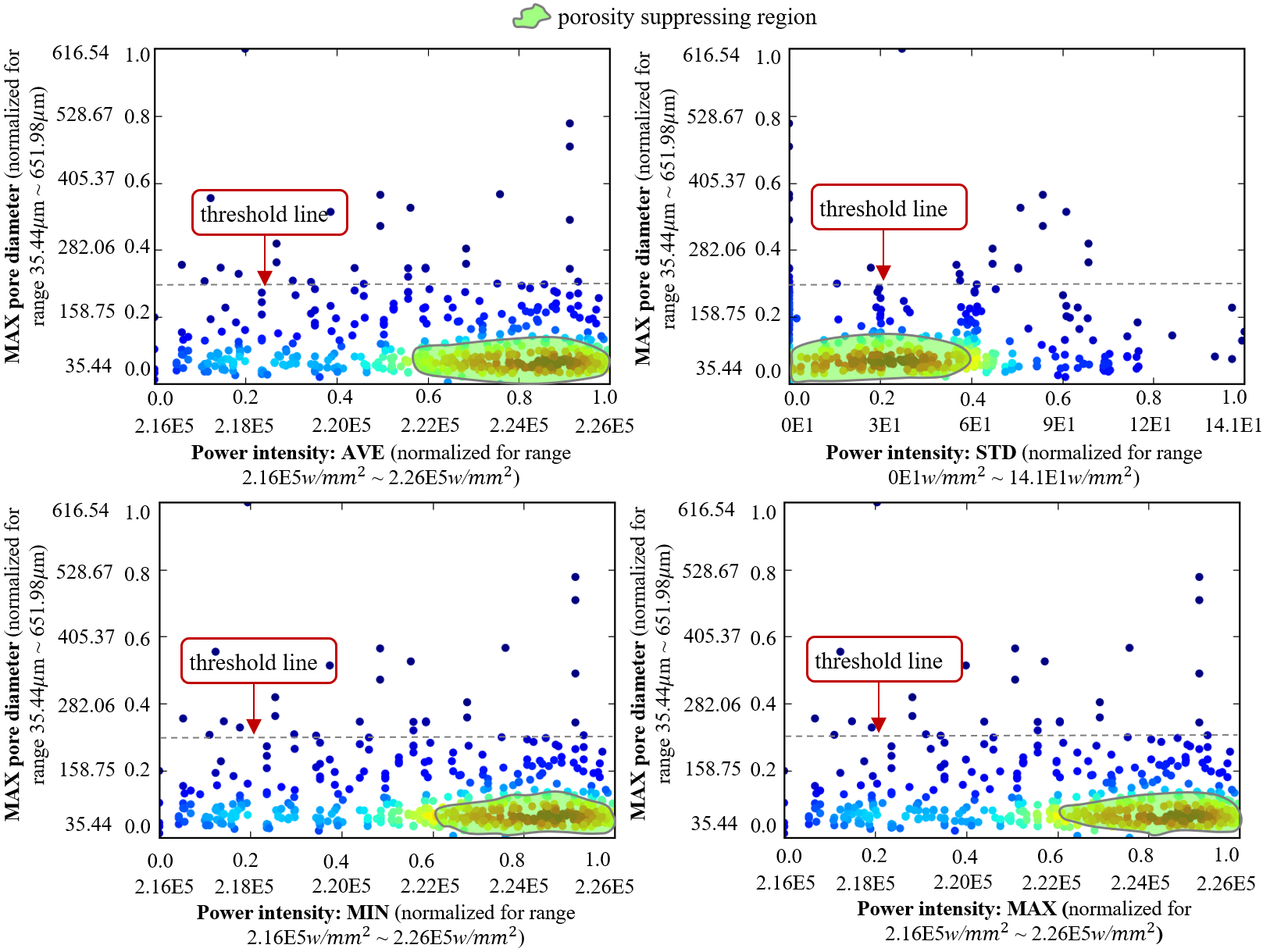}
\caption{\ch{Sensitive ranges of the physical effect {power intensity} for pore generation: maximum diameter of the generated pores.}}
\label{fig:14}
\end{figure*}

\begin{figure*}[ht]
\centering
\includegraphics[width=0.9 \linewidth ]{ 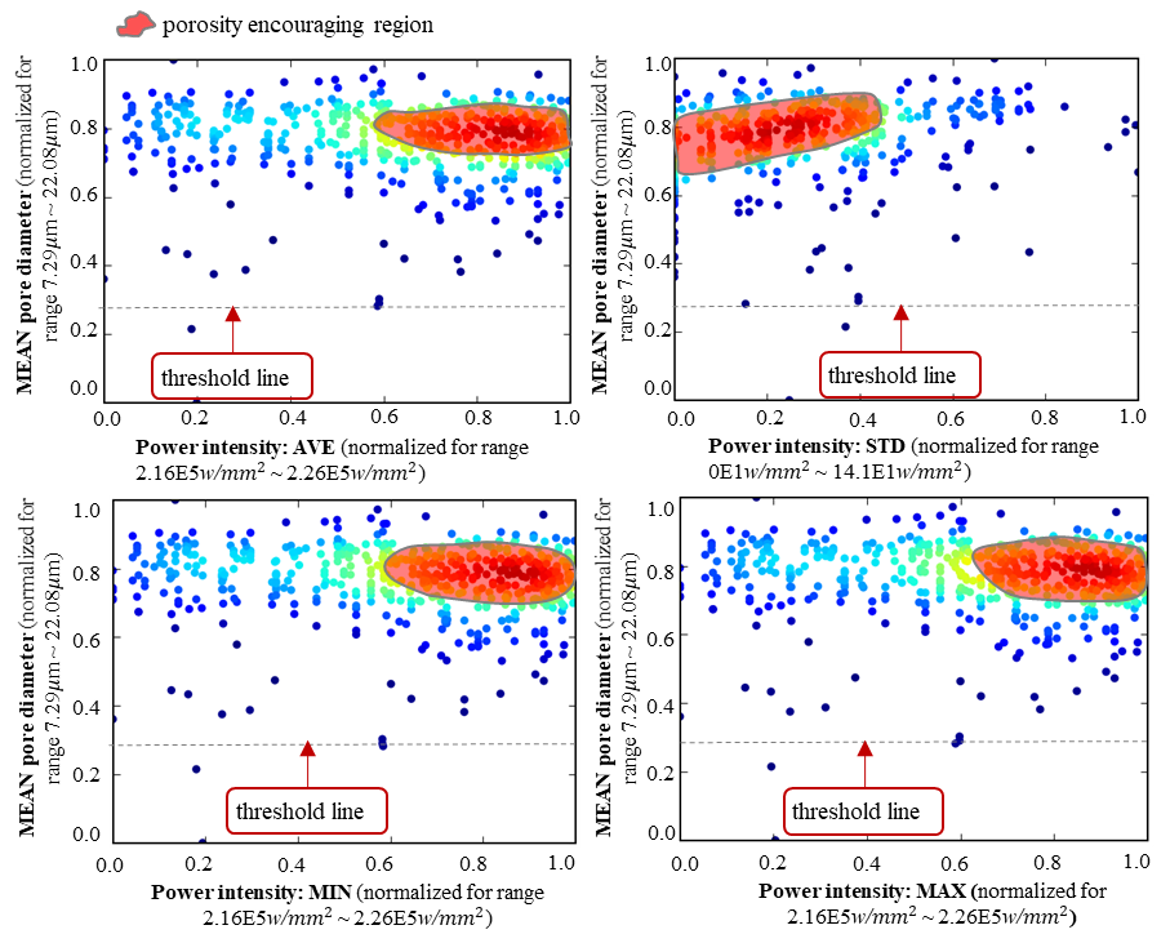}
\caption{Sensitive ranges of the physical effect {power intensity} for pore generation: mean diameter of the generated pores.}
\label{fig:15}
\end{figure*}

Notice that porosity predictions for mean/median pore diameter and pore spacing are generally good, with average prediction error about 12\% while porosity prediction for maximum pore diameter is bad with error at about 27\%. This is because mean/median pore diameter and median pore spacing are relatively stable while maximum pore diameter is not stable, caused by uncertain and complex factors during system setting, sample preparation and characterization process. \ch{Generally, these factors could be the following: (1) extra supporting material attachments and part deformation in the part removing stage, resulting large-diameter pores in the sample-plate interface; (2) Varying sweeper stress as part pose varies, generating large-diameter pores; (3) uncertain powder interactions as powder distributions vary, generating large-diameter pores; and (4) Even though laser-related influence is playing a primary role, other factors, such as powder size, and blade motion direction and speed, are also influential, resulting in some outliers given this laser-based porosity prediction model.} Even though prediction of maximum porosity is with relatively low accuracy in the current research stage, it is still important to investigate the maximum pore generation mechanism, so that the maximum pore diameter closely influences part mechanical properties such as strength, stiffness, and conductivity. The physics informed model has the lowest error in predicting the maximum pore size, indicating this model is useful in defect prediction.

To compare algorithm performances, the prediction errors are averaged from model performances based on {pass part, flag part, fail part} and four types of porosity levels. As it shown in Fig. \ref{fig:12}, the six algorithms {Linear, GPR, svrL, svrG, svrRBF, svrP} have averaged prediction errors of 17\%, 15\%, 15\%, 13\%, 13\%, 16\%, respectively. \ch{The algorithms svrG and svrRBF show best performance for our case study on physics-informed porosity analysis.}

To go beyond black-box porosity prediction and investigate the data quality influence on model performances, correlations between real porosities and predicted porosities are visualized, shown in Fig. \ref{fig:13}. To maximally reduce the algorithms’ influence towards physics-porosity correlation, the best-performed model, svrRBF, was selected for data predictability analysis. Prediction errors are averaged from the performances of three quality-level models. Each row (three figures) is one of the four porosity types {MAX, MEAN, MEDIAN, SPACING}, and each column (four figures) is one of the three prediction models {setting model, combined model, physics-informed model}. As (a)-(c) show, about 67\% maximum porosity values are concentrated at good-quality region [35.44 $\mu m$, 97.10 $\mu m$]. Starting from the fair region ($>$97.10 $\mu m$), prediction errors increase to beyond 20\%. Starting from about 341.71 $\mu m$, data becomes more unpredictable, with large error increasing. For mean and median pore diameter,  as shown in (d)-(i) , about 80\% porosity values are concentrated in range of [16 $\mu m$, 20 $\mu m$], with small error ($<$20\%) and strong predictability; for porosity values decreasing from 15.00 $\mu m$, error begins to largely increase, leading to prediction performance derivation. As (j)-(l) show, more than 80\% porosity values are concentrated in range [54.87 $\mu m$, 63.78 $\mu m$], with smaller error ($<$20\%) and strong predictability; for values out of this range, error begins to largely increase, leading to model performance derivation.

\subsection{Pore generation explanation}
To identify pore generation regions during part manufacturing, impacts of physical effects were investigated by a novel physics-informed pore generation explanation (PIPGE) method. The PIPGE method visualizes the relationship between the physics effects and porosity via physics-porosity maps, as shown in Figs. \ref{fig:14} and \ref{fig:15}. \ch{From a statistical perspective, influenced regions in physics-porosity maps are summarized as two types: (1) Porosity suppressing region in Fig. \ref{fig:14} and (2) Porosity encouraging region in Fig. \ref{fig:15}. These regions are defined as physics-effect parameter range at the cluster of lower (suppressing) or higher (encouraging) porosity values in the physics-porosity map.} In the encouraging region, the porosity level is higher than 0.3 in a normalized porosity range. The corresponding physical effect range of this region is encouraging generation of large pores, and the physical effects in the encouraging region are highly likely to lead an undesired high porosity level for a part. \ch{In the suppressing region as shown in Fig. 14, the porosity level is lower than 0.5 in a normalized porosity range, and the sample density is relatively high. The maximum pore size larger than 0.3, being about 220 $\mu m$, could be considered as outliers. These large size pores usually exist at sample-substrate plate interface in the sample removing or preparation process as we discussed before. The same threshold line 0.3 is set for porosity encouraging region in Fig. 15. } The corresponding physical effect range of this region is suppressing generation of large pores, and the physical effects in the suppressing region are highly likely to lead a favorable low porosity level for a part. \ch{For good quality control, the pore suppressing region could be achieved by adjusting machine-settings, and porosity encouraging regions can be avoided as much as possible. }

Each dot in a physics-porosity map denotes a specific (physical-effect vs. porosity-type) correlation. To make the visualizations of effect-porosity correlations are understandable, physical effect values were normalized by Max-Min scaling, in which the maximum value is normalized as 1, the minimum value is normalized as 0, and other values are linearly scaled accordingly. Real physical effect values are in the brackets after the Max-Min normalized values. Limited by the length of this paper, only the (power-intensity vs. maximum-pore-diameter) correlation map is shown in Fig. \ref{fig:14}. Other physics-porosity maps can be found in the appendix. As shown in Fig. \ref{fig:14}, typical suppressing regions of power intensity (unit is $w/mm^2$) for max pore diameter are estimated as AVE[0.58E5, 1.00E5], SDEV[0.00 E1, 0.40 E1], MIN[0.60 E5, 1.00 E5], and MAX[0.60E5, 1.00E5]; typical encouraging regions are not obviously observed. With these regions, the generation of maximum pore sizes is mainly caused by low to average power intensity (lower than AVE 0.58. Please see detailed real value in the attachment file), high laser energy variance ($>$STD 0.4), and wide max-min range (beyond MIN [0.6,1.00]). This also tells in order to suppress pore generation the machine setting should be adjusted to make energy intensity fall into the typical suppressing regions.

By using the same intuition introduced above, influence regions for other types of physics-porosity maps between three physical effects and four porosity levels {maximum pore, mean pore, median pore, median pore spacing} were analyzed. Full ranges for real values for physical effects are listed in the attachment file.

\section{Conclusion \& Future work}
\ch{A physics-informed data-driven model for predicting and explaining pore generation in laser powder bed fusion metal AM has been presented in this paper. The model prediction accuracy and analysis on porosity were studied based on Inconel 718 parts printed on Concept Laser M2 Dual-Laser Cusing machine. The comparable performances of the physics-informed model and the machine-setting model have proven the feasibility and effectiveness of using physical effects to predict part porosity. Reasonable interpretations of physical effects’ influence towards pore generation, such as suppressing and encouraging influence, and physical effect sensitive ranges, suggested that the physics-informed model can give insights on pore generation region. With a physics-perspective understanding of porosity-machine setting correlation, this physics-informed model holds a potential in reverse design of process parameters such as laser power and scan speed for part quality requirements. Although in this work we only validated that the physics-informed ML models can work on Concept Laser M2 machine, the method we developed for physics effects-based modeling has the potential to generalize to other laser bed fusion-based 3D printing machines, such as Concept Laser M/M3, EOS M270/280/290, and Prox DMP 300/320. The physics model uses machine-independent parameters. If machine settings in various manufacturing scenarios could be comprehensively interpreted into physics effects parameters, a more accurate and computational efficient physics-based data-driven ML model would be achieved.}

\ch{In the future, the following work could be done: (1) More physical effects such as temperature and material energy absorption will be involved to enrich physical effects, further improving model performance and validating the model flexibility. The mechanism of pore generation will be described with more comprehensive, informative physics effects in the laser-based manufacturing process. (2) More properties of printed part will be experimentally measured with mechanical tests. Currently the developed physics-informed ML model works well on porosity analysis. Since some mechanical properties such as yield strength and strain are closely related with part porosity levels. The prediction feasibility of model on these mechanical properties are expected as well and will be evaluated. (3) The physics-informed ML model will be developed for evaluation and validation on different machines such as Prox DMP 320A and EOS M290. Specifically, these machines are similar in physics effects we generated for laser powder bed fusion AM technology, but the machines in industry are often come from different manufacturer and use different technologies. For example, our case study Concept Laser M2 installs a dual-laser system with 200 W for each. The Prox DMP 320A from 3D systems company installs a fiber laser in the power range up to 500 W. The build envelope 275 × 275 × 420 mm of Prox DMP 320A is also large than Concept Laser M2. The evaluation of developed method on these kinds of different scenarios will provide the evidences of the methods’ flexible and generalizable and increase its impact on industries.}

\begin{acknowledgements}
Thanks for valuable suggestions and experiment assistance from Dr. Aaron Stebner, Dr.Branden B. Kappes, Mr. Henry Geerlings, Mr. Senthamilaruvi Moorthy.   
\end{acknowledgements}

%
%


\begin{thebibliography}{}
%
%
\bibitem{ref1}
Bi G, Sun CN, Gasser A (2013) Study on influential factors for process monitoring and control in laser aided additive manufacturing. J Mater Process Technol 213:463–468

\bibitem{ref2}
Mellor S, Hao L, Zhang D (2014) Additive manufacturing: A framework for implementation. Int J Prod Econ 149:194–201

\bibitem{ref3}
Huang R, Riddle M, Graziano D, et al (2016) Energy and emissions saving potential of additive manufacturing: the case of lightweight aircraft components. J Clean Prod 135:1559–1570

\bibitem{ref4}
Benesty J, Chen J, Huang Y, Cohen I (2009) Pearson correlation coefficient. In: Noise reduction in speech processing. Springer, pp 1–4

\bibitem{ref5}
Frazier WE (2014) Metal additive manufacturing: a review. J Mater Eng Perform 23:1917–1928

\bibitem{ref6}
Tapia G, Elwany A (2014) A review on process monitoring and control in metal-based additive manufacturing. J Manuf Sci Eng 136:

\bibitem{ref7}
Vilaro T, Colin C, Bartout J-D, et al (2012) Microstructural and mechanical approaches of the selective laser melting process applied to a nickel-base superalloy. Mater Sci Eng A 534:446–451

\bibitem{ref8}
Thijs L, Verhaeghe F, Craeghs T, et al (2010) A study of the microstructural evolution during selective laser melting of Ti–6Al–4V. Acta Mater 58:3303–3312

\bibitem{ref9}
Wits WW, Carmignato S, Zanini F, Vaneker THJ (2016) Porosity testing methods for the quality assessment of selective laser melted parts. CIRP Ann 65:201–204

\bibitem{ref10}
Gong H, Rafi K, Gu H, et al (2015) Influence of defects on mechanical properties of Ti–6Al–4 V components produced by selective laser melting and electron beam melting. Mater Des 86:545–554

\bibitem{ref11}
Tolochko NK, Mozzharov SE, Yadroitsev IA, et al (2004) Balling processes during selective laser treatment of powders. Rapid Prototyp J

\bibitem{ref12}
Oliveira JP, LaLonde AD, Ma J (2020) Processing parameters in laser powder bed fusion metal additive manufacturing. Mater Des 193:108762

\bibitem{ref13}
Oliveira JP, Santos TG, Miranda RM (2020) Revisiting fundamental welding concepts to improve additive manufacturing: From theory to practice. Prog Mater Sci 107:100590

\bibitem{ref14}
Slotwinski JA, Garboczi EJ, Hebenstreit KM (2014) Porosity measurements and analysis for metal additive manufacturing process control. J Res Natl Inst Stand Technol 119:494

\bibitem{ref15}
Lewis G (2013) Properties of open-cell porous metals and alloys for orthopaedic applications. J Mater Sci Mater Med 24:2293–2325

\bibitem{ref16}
Vaezi M, Seitz H, Yang S (2013) A review on 3D micro-additive manufacturing technologies. Int J Adv Manuf Technol 67:1721–1754

\bibitem{ref17}
Song B, Dong S, Zhang B, et al (2012) Effects of processing parameters on microstructure and mechanical property of selective laser melted Ti6Al4V. Mater Des 35:120–125

\bibitem{ref18}
Zhang S, Wei Q, Cheng L, et al (2014) Effects of scan line spacing on pore characteristics and mechanical properties of porous Ti6Al4V implants fabricated by selective laser melting. Mater Des 63:185–193

\bibitem{ref19}
Xu W, Lui EW, Pateras A, et al (2017) In situ tailoring microstructure in additively manufactured Ti-6Al-4V for superior mechanical performance. Acta Mater 125:390–400

\bibitem{ref20}
King W, Anderson AT, Ferencz RM, et al (2015) Overview of modelling and simulation of metal powder bed fusion process at Lawrence Livermore National Laboratory. Mater Sci Technol 31:957–968


\bibitem{ref21}
Wu Y-C, San C-H, Chang C-H, et al (2018) Numerical modeling of melt-pool behavior in selective laser melting with random powder distribution and experimental validation. J Mater Process Technol 254:72–78

\bibitem{ref22}	
Patil N, Pal D, Stucker B (2013) A new finite element solver using numerical eigen modes for fast simulation of additive manufacturing processes. In: Proceedings of the Solid Freeform Fabrication Symposium. pp 12–14

\bibitem{ref23}
Li S, Xiao H, Liu K, et al (2017) Melt-pool motion, temperature variation and dendritic morphology of Inconel 718 during pulsed-and continuous-wave laser additive manufacturing: A comparative study. Mater Des 119:351–360

\bibitem{ref24}
Ye J, Khairallah SA, Rubenchik AM, et al (2019) Energy coupling mechanisms and scaling behavior associated with laser powder bed fusion additive manufacturing. Adv Eng Mater 21:1900185

\bibitem{ref25}
Gong H, Rafi K, Gu H, et al (2014) Analysis of defect generation in Ti–6Al–4V parts made using powder bed fusion additive manufacturing processes. Addit Manuf 1:87–98

\bibitem{ref26}
Khademzadeh S, Carmignato S, Parvin N, et al (2016) Micro porosity analysis in additive manufactured NiTi parts using micro computed tomography and electron microscopy. Mater Des 90:745–752

\bibitem{ref27}
Aboulkhair NT, Everitt NM, Ashcroft I, Tuck C (2014) Reducing porosity in AlSi10Mg parts processed by selective laser melting. Addit Manuf 1:77–86

\bibitem{ref28}
Johnson NS, Vulimiri PS, To AC, et al (2020) Machine Learning for Materials Developments in Metals Additive Manufacturing. arXiv Prepr arXiv200505235

\bibitem{ref29}
Gobert C, Reutzel EW, Petrich J, et al (2018) Application of supervised machine learning for defect detection during metallic powder bed fusion additive manufacturing using high resolution imaging. Addit Manuf 21:517–528

\bibitem{ref30}
Zhang W, Mehta A, Desai PS, Higgs C (2017) Machine learning enabled powder spreading process map for metal additive manufacturing (AM). In: Int. Solid Free Form Fabr. Symp. Austin, TX. pp 1235–1249

\bibitem{ref31}
Scime L, Beuth J (2019) Using machine learning to identify in-situ melt pool signatures indicative of flaw formation in a laser powder bed fusion additive manufacturing process. Addit Manuf 25:151–165

\bibitem{ref32}
Donegan SP, Schwalbach EJ, Groeber MA (2020) Zoning additive manufacturing process histories using unsupervised machine learning. Mater Charact 161:110123

\bibitem{ref33}
Wang C, Tan XP, Tor SB, Lim CS (2020) Machine learning in additive manufacturing: State-of-the-art and perspectives. Addit Manuf 101538

\bibitem{ref34}
Kappes B, Moorthy S, Drake D, et al (2018) Machine learning to optimize additive manufacturing parameters for laser powder bed fusion of Inconel 718. In: Proceedings of the 9th International Symposium on Superalloy 718 \& Derivatives: Energy, Aerospace, and Industrial Applications. Springer, pp 595–610

\bibitem{ref35}
Liu, S; Stebner Aaron P; Kappes Branden B; Zhang X (2020) Machine Learning for Knowledge Transfer Across Multiple Metals Additive Manufacturing Printers. arXiv e-prints

\bibitem{ref36}
Tapia G, Elwany AH, Sang H (2016) Prediction of porosity in metal-based additive manufacturing using spatial Gaussian process models. Addit Manuf 12:282–290

\bibitem{ref37}
Kamath C, El-Dasher B, Gallegos GF, et al (2014) Density of additively-manufactured, 316L SS parts using laser powder-bed fusion at powers up to 400 W. Int J Adv Manuf Technol 74:65–78

\bibitem{ref38}
Garg A, Lam JSL, Savalani MM (2015) A new computational intelligence approach in formulation of functional relationship of open porosity of the additive manufacturing process. Int J Adv Manuf Technol 80:555–565

\bibitem{ref39}
Shamsaei N, Yadollahi A, Bian L, Thompson SM (2015) An overview of Direct Laser Deposition for additive manufacturing; Part II: Mechanical behavior, process parameter optimization and control. Addit Manuf 8:12–35

\bibitem{ref40}
Liu S, Kappes BB, Amin-ahmadi B, et al (2020) A physics-informed feature engineering approach to use machine learning with limited amounts of data for alloy design: shape memory alloy demonstration. arXiv Prepr arXiv200301878

\bibitem{ref41}
Sangid MD (2020) Coupling in situ experiments and modeling–Opportunities for data fusion, machine learning, and discovery of emergent behavior. Curr Opin Solid State Mater Sci 24:100797

\bibitem{ref42}
Wang Z, Liu P, Ji Y, et al (2019) Uncertainty quantification in metallic additive manufacturing through physics-informed data-driven modeling. JOM 71:2625–2634

\bibitem{ref43}
Yan W, Lin S, Kafka OL, et al (2018) Data-driven multi-scale multi-physics models to derive process–structure–property relationships for additive manufacturing. Comput Mech 61:521–541

\bibitem{ref44}
Zhu Q, Liu Z, Yan J (2020) Machine learning for metal additive manufacturing: Predicting temperature and melt pool fluid dynamics using physics-informed neural networks. arXiv Prepr arXiv200813547

\bibitem{ref45}
Ren F, Ward L, Williams T, et al (2018) Accelerated discovery of metallic glasses through iteration of machine learning and high-throughput experiments. Sci Adv 4:eaaq1566

\bibitem{ref46}
Ward L, Agrawal A, Choudhary A, Wolverton C (2016) A general-purpose machine learning framework for predicting properties of inorganic materials. npj Comput Mater 2:16028

\bibitem{ref47}
Ghiringhelli LM, Vybiral J, Levchenko S V, et al (2015) Big data of materials science: critical role of the descriptor. Phys Rev Lett 114:105503

\bibitem{ref48}
Stanev V, Oses C, Kusne AG, et al (2018) Machine learning modeling of superconducting critical temperature. npj Comput Mater 4:1–14

\bibitem{ref49}
Geerlings H (2018) TRACR: a software pipeline for high-throughput materials image analysis--an additive manufacturing study

\bibitem{ref50}
Nichols EF, Hull GF (1903) The pressure due to radiation.(second paper.). Phys Rev (Series I) 17:26

\bibitem{ref51}
Hecht E (1998) Optics 4th edition. Optics. Addison Wesley Longman Inc 1:629

\bibitem{ref52}
Balachandran P V, Xue D, Theiler J, et al (2016) Adaptive strategies for materials design using uncertainties. Sci Rep 6:19660

\bibitem{ref53}
Williams CKI, Rasmussen CE (1996) Gaussian processes for regression. In: Advances in neural information processing systems. pp 514–520

\bibitem{ref54}
Bishop CM (2006) Pattern recognition and machine learning. springer

\bibitem{ref55}
Cawley GC, Talbot NLC (2003) Efficient leave-one-out cross-validation of kernel fisher discriminant classifiers. Pattern Recognit 36:2585–2592


\end{thebibliography}


\section{Attachments}

\ch{\begin{enumerate}
\item Visualization for porosity distribution and physics effects
\item Sensitive ranges and real-value range for physics
\item Physics-porosity correlation
\item Physics effects visualization
\end{enumerate}}

\end{document}